%% file: faceMirrorSymmetry_mainVer22_SIver46_memo_ver01.tex
\documentclass[11pt,a4paper]{article} 

\usepackage[utf8]{inputenc}
\usepackage{multicol}
\usepackage{verbatim}

\usepackage{subfigure}
\usepackage[numbers]{natbib}

\usepackage{bm}
\usepackage{amsmath}

\usepackage{boxedminipage}

\usepackage{listings}

\usepackage{ifpdf}

\usepackage{tgpagella}
\usepackage[T1]{fontenc}
\usepackage{fancyhdr}   

\usepackage[svgnames]{xcolor}
\usepackage{xcolor,colortbl}  
\usepackage[margin= 0.8in,top=12mm]{geometry}        
\newcommand*{\titleAT}{\begingroup
  \newlength{\drop}
  \drop=0.04\textheight  
  \includegraphics[scale=1.5]{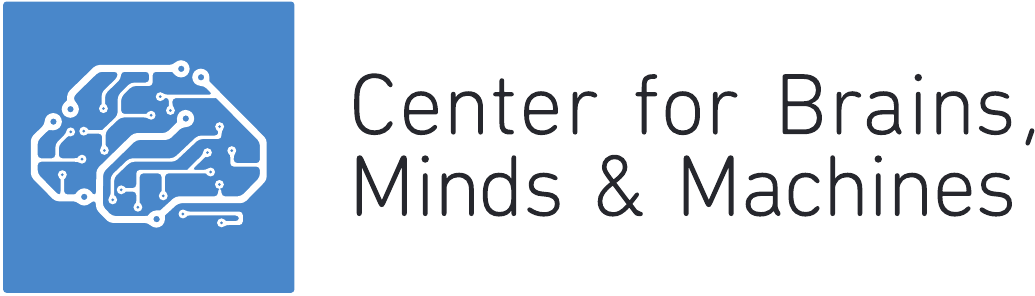} 

  \textcolor{CornflowerBlue}{\rule{\textwidth}{3 pt}} 
  \par   

  \vspace{2pt}\vspace{-\baselineskip} 
  \rule{\textwidth}{0.4pt} 
  \par 

  \vspace{\drop} 
  \textbf{\large{CBMM Memo No. \memonumber}}   \hfill    \textbf{\large{\memodate}} 
  \vspace{\drop} 
  \begin{center}
    \textbf{\huge{\memotitle}}\\
    \vspace{0.4\drop} 
    \textbf{\Large{by}}\\
    \vspace{0.4\drop} 
    \large{\memoauthors}
  \end{center}
  \textbf{\large{\noindent Abstract}:} {\memoabstract}
 
  \textcolor{CornflowerBlue}{\rule{\textwidth}{3 pt}}\par
  \vspace{2pt}\vspace{-\baselineskip}
  \rule{\textwidth}{0.4pt}\par  
  
  \begin{minipage}{.15\linewidth}
    \includegraphics[scale=0.1]{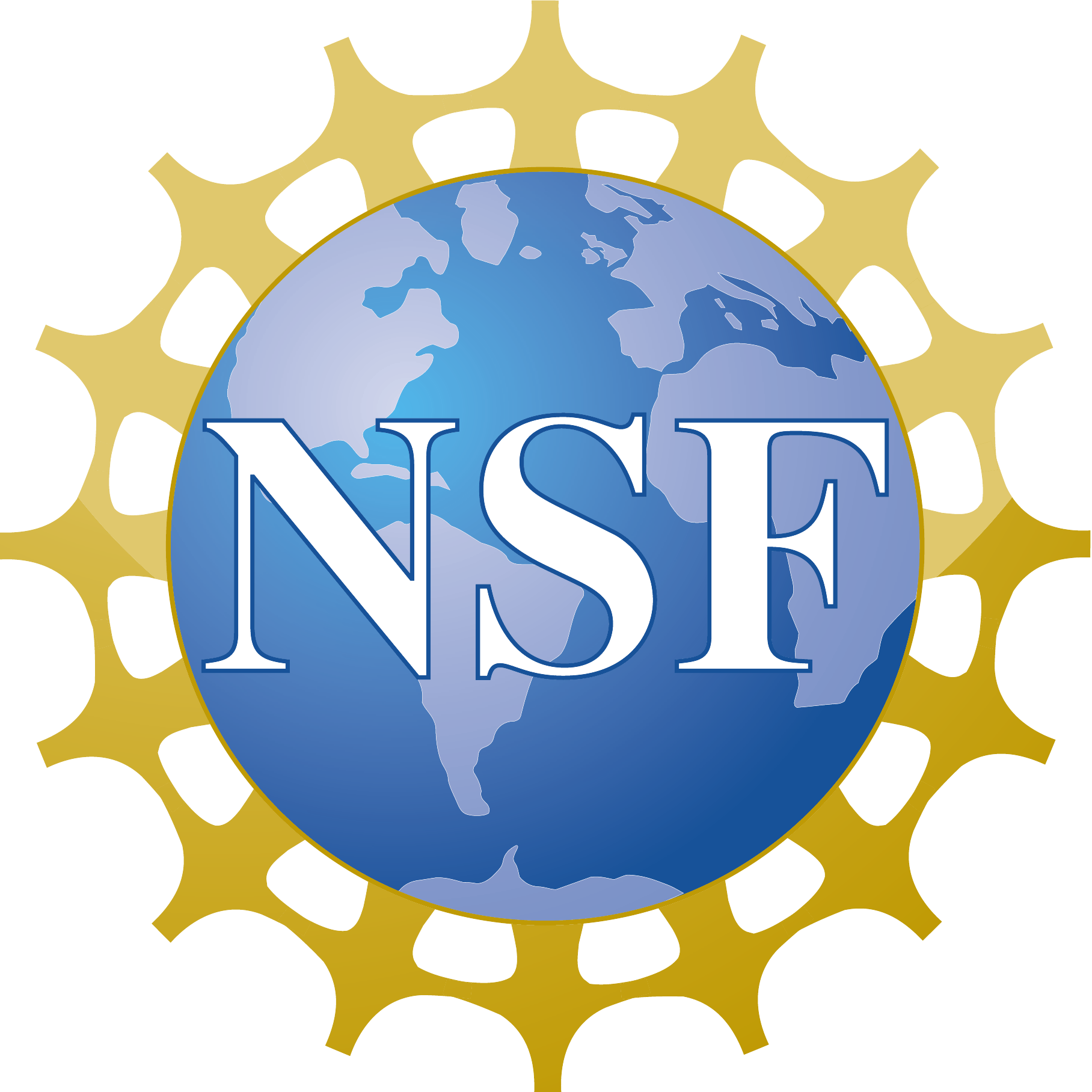}
  \end{minipage}
  \begin{minipage}{.84\linewidth}
    \textbf{\large{This work was supported by the Center for Brains, Minds and Machines (CBMM), funded by NSF STC award  CCF - 1231216.}}
  \end{minipage}
  \endgroup} 

\ifpdf
\usepackage[pdftex]{graphicx}
\else
\usepackage{graphicx}
\fi
\usepackage{rotating}
\usepackage{blindtext} 
\ifnum\pdfoutput>0			
\usepackage[pdftex, 			
pagebackref = false, 			
bookmarks = true, 			
bookmarksnumbered = false, 	
hyperindex = true,			
linktocpage = true,			
pdfpagemode = UseNone, 		
pdfstartview = Fit, 			
pdfpagelayout = OneColumn,
colorlinks = true, 			
urlcolor = blue, 				
citecolor = blue,
pdfborder = {0 0 0} 			
]{hyperref} 					
\fi

\ifpdf
\DeclareGraphicsExtensions{.png, .pdf, .jpg, .tif}
\else
\DeclareGraphicsExtensions{.eps, .jpg}
\fi

\input{macros.tex}

\providecommand{\scal}[2]{\left\langle{#1},{#2}\right\rangle} 
\usepackage[scaled]{helvet}
\providecommand{\nor}[1]{\left\lVert {#1} \right\rVert}
\providecommand{\scal}[2]{\left\langle{#1},{#2}\right\rangle}

\newcommand{\T}{\mathcal T}

\newtheorem{theorem}{Theorem}
\newtheorem{definition*}{Definition}

\begin{document}

\def\memonumber{049} 
\def\memodate{\today}      
\def\memotitle{View-tolerant face recognition and Hebbian learning imply mirror-symmetric neural tuning to head orientation}
\def\memoauthors{Joel Z. Leibo$^{1}$, Qianli Liao$^{1}$, Winrich Freiwald$^{1, 2}$, Fabio Anselmi$^{1, 3}$, Tomaso Poggio$^{1}$ \\
\small 
  1: Center for Brains, Minds, and Machines and McGovern Institute for Brain Research at MIT, Cambridge, MA, USA \\
  2: Laboratory of Neural Systems, The Rockefeller University, New York, NY, USA \\
  3: Istituto Italiano di Tecnologia, Genova, Italy \\
\normalsize  
}

\def\memoabstract{
The primate brain contains a hierarchy of visual
  areas, dubbed the ventral stream, which rapidly computes object
  representations that are both specific for object identity and
  relatively robust against identity-preserving transformations like
  depth-rotations \citep{Logothetis1995, logothetis1996visual,
    Hung2005, DiCarlo2012}. Current computational models of object
  recognition, including recent deep learning networks, generate these
  properties through a hierarchy of alternating selectivity-increasing
  filtering and tolerance-increasing pooling operations, similar to
  simple-complex cells operations \citep{Serre2007a, bart2008class,
    Rolls2012, leibo2015invariance}. While simulations of these models
  recapitulate the ventral stream's progression from early
  view-specific to late view-tolerant representations, they fail to generate the most
  salient property of the intermediate representation for faces found
  in the brain: mirror-symmetric tuning of the neural population to
  head orientation \citep{freiwald2010functional}. Here we prove that a
  class of hierarchical architectures and a broad set of biologically
  plausible learning rules can provide approximate invariance at the
  top level of the network. While most of the learning rules do not
  yield mirror-symmetry in the mid-level representations, we
  characterize a specific biologically-plausible Hebb-type learning
  rule that is guaranteed to generate mirror-symmetric tuning to faces tuning
  at intermediate levels of the architecture.
} 

\titleAT
\newpage

\begin{figure}[h]  
\begin{center}
\centerline{\includegraphics[width=0.6\textwidth]{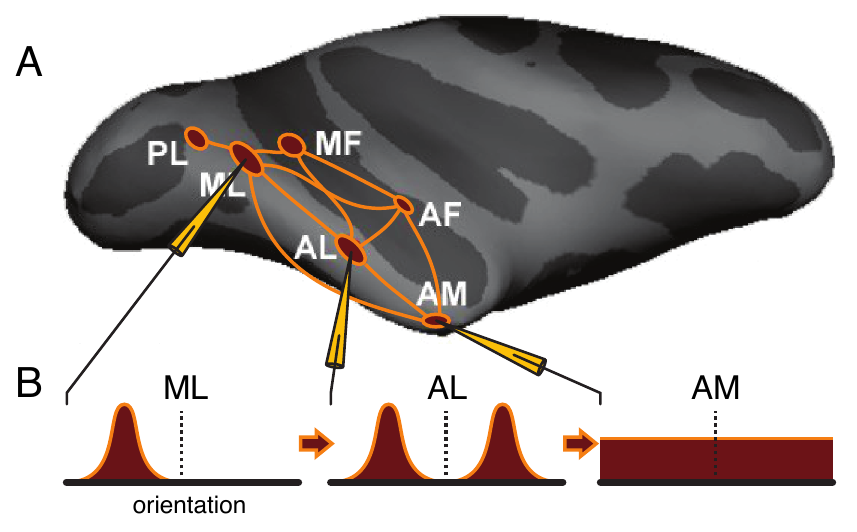}}    
\caption{Schematic of the macaque face-patch system \citep{Moeller2008,
    Tsao2008, freiwald2010functional}. (A) Side view of
  computer-inflated macaque cortex with six areas of face-selective
  cortex (red) in the temporal lobe together with connectivity graph
  (orange). Face areas are named based on their anatomical location:
  PL, posterior lateral; ML; middle lateral; MF, middle fundus; AL,
  anterior lateral; AF, anterior fundus ; AM, anterior medial (3), and
  have been found to be directly connected to each other to form a
  face-processing network \citep{Moeller2008}. Recordings from three
  face areas, ML, AL, AM, during presentations of faces at different
  head orientations revealed qualitatively different tuning
  properties, schematized in B. (B) Prototypical ML neurons are tuned
  to head orientation, e.g., as shown, a left profile. A prototypical
  neuron in AL, when tuned to one profile view, is tuned to the
  mirror-symmetric profile view as well. And a typical neuron in AM is
  only weakly tuned to head orientation. Because of this increasing
  invariance to in-depth rotation, increasing to invariance to size
  and position (not shown) and increased average response latencies
  from ML to AL to AM, it is thought that the main AL properties,
  including mirror-symmetry, have to be understood as transformations
  of ML representations, and the main AM properties as transformations
  of AL representations
  \citep{freiwald2010functional}.\label{fig:face_patches}}
\end{center}
\end{figure}

The ventral stream rapidly computes image representations that are
simultaneously tolerant of identity-preserving transformations and
discriminative enough to support robust recognition. The ventral
stream of the macaque brain contains discrete patches of cortex that
support the processing of images of faces \citep{Tsao2003, Tsao2006,
  ku2011fmri, afraz2015optogenetic}. Face patches are selectively
interconnected to form a face-processing network
\citep{Moeller2008}. Face patches are arranged along an
occipito-temporal axis (from the middle lateral (ML) and middle fundus (MF)
patches, through the antero-lateral face patch (AL), and culminating in the antero-medial (AM) patch
\citep{tsao2008comparing} (Fig. \ref{fig:face_patches}-A)) along which
response latencies increase systematically from ML/MF via AL to AM,
suggesting sequential forward-processing \citep{freiwald2010functional}.

Face patches differ qualitatively in how they represent identity
across head orientations \citep{freiwald2010functional}. Neurons in the
ML/MF patches are view-specific, while neurons in AM approach
view-invariance. Furthermore, spatial position and size invariance
increase from ML/MF to AL, and further to AM
\citep{freiwald2010functional}. These properties of the face-processing
network replicate the general trend of the ventral stream as
summarized in \citep{logothetis1996visual, Riesenhuber1999,
  DiCarlo2012} and conform to the concept of a feedforward processing
hierarchy.

Several hierarchical models of object recognition
\citep{Fukushima1980,Poggio1990a, Riesenhuber1999, Bart2004} and face
recognition \citep{bart2008class, leibo2015invariance,
  farzmahdi2016specialized} feature a progression from view-specific
early processing stages to view-invariant later processing stages
similar to ML/MF and AM, respectively. Simulations have shown that
view-based models can achieve an AM-like representation by successively
pooling the responses of view-tuned units like those found in the
early processing stage ML/MF. The theoretical underpinnings of this
property are described in the Appendix (section 1.1).

Neurons in the intermediate face area AL, but not in preceding areas
ML/MF, exhibit mirror-symmetric head orientation tuning 
\citep{freiwald2010functional}. That is, an AL neuron tuned to one
profile view of the head typically responds similarly to the opposite
profile, but not to the front view
(Fig. \ref{fig:face_patches}-B). This phenomenon is not predicted by
simulations of classical and current view-based computational models
of the ventral stream \citep{YaminsDicarlo2016}. In this paper we
ask why  the primate brain may compute a mirror symmetric
representation as a necessary intermediate step towards invariant
face-representation and what this tells us about the brain's mechanisms
of learning.

\section*{Results}

\begin{figure}
\begin{center}
\centerline{\includegraphics[width=0.8\textwidth]{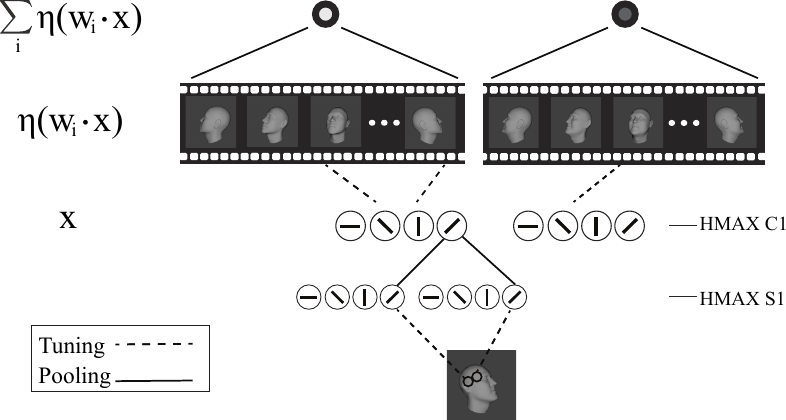}} 
\caption{Illustration of the model. Inputs are encoded in HMAX C1 \citep{Riesenhuber1999}, then
  projected onto $w_i$. In the view-based model the $w_i$ represent faces at specific views. In the Oja-model, the $w_i$ are principal components. Units in the output layer pool over all the units in the previous layer corresponding to projections onto the same template individual's views (view-based model) or PCs (Oja-model).\label{fig:schematic}}
\end{center}
\end{figure}

\subsection*{Assumptions underlying the model}

We consider a feedforward face-processing hierarchy as a model
for how the ventral stream rapidly computes invariant
representations. Invariant information can be decoded from
inferotemporal cortex, and the face areas within it, roughly 100ms
after stimulus presentation \citep{Hung2005,
  meyers2015intelligent}. This is too fast of a timescale for feedback
to play a large role \citep{Hung2005, Thorpe1996,
  isik2014dynamics}. Thus while the actual face processing system
might operate in other modes as well, all indications are that
fundamental properties of shape-selectivity and invariance need to be
explained as a property of feedforward processing.

The population of neurons in ML/MF is highly face selective
\citep{Tsao2006} and incoming information can be thought of as passing
through a face-likeness filter. We thus assume the existence of a
  functional gate that routes only images of face-like objects at the
input of the face system. The existence of large ``face-like''
templates or filters explains many of the so-called holistic effects
of face perception, including face inversion and the composite face
\citep{young1987configurational} effect \citep{Tan2016, farzmahdi2016specialized}. This property has one further
computational implication: it provides an automatic face-specific
gating mechanism to the face-processing system.

We make the standard assumption that a neuron's basic operation is a pooled dot product between inputs $x$ and synaptic weight vectors $\{w_i\}$, yielding
complex-like cells as
 
\begin{equation}
\label{signature}
\mu^k(x) = \frac{1}{|G|} \sum_{i=1}^{|G|}  \eta(\scal{x}{g_i w^k})
\end{equation}

\noindent where $\eta: \R \rightarrow \R$ is a nonlinear function e.g., squaring as in \citep{adelson1985spatiotemporal}.  We suppose that $g_i \in G$ are image plane transformations corresponding to rotations in depth of the face. Note that $G$ is a set of transformations but it is not a group (see Appendix section 1.1). We call $\vec{\mu}(x)\in \R^{K}$ the signature of image $x$. 

\subsection*{Approximate view invariance}

The model of Eq. \eqref{signature} encodes a
novel face by its similarity to a set of stored template faces. For
example, the $g_i w^k$ could correspond to views $i$ of each of a set
of well-known individuals $k$ from an early developmental period e.g.,
parents, caretakers, etc. One could regard the acquisition of this set
of familiar faces as the algorithm's (unsupervised) training phase. To
see why the algorithm works, consider that whenever $w^k_i$ encodes a
non-matching orientation to $I$, the value of $\scal{x}{g_i w^k}$ will
be very low. Among the $w^k$ tuned to the correct orientation, there
will be a range of response values since different template faces will
have different levels of similarity to $I$. When the novel face
appears at a different orientation, the only effect is to change which
specific view-tuned units carry its signature. Since the pooled neural
response is computed by summing over these, the large responses
carrying the signature will dominate. Thus the pooled neural response
will be approximatively unchanged by rotation (see the Appendix section
1).  Since these models are based on stored associations of frames, they
can be interpreted as taking advantage of temporal continuity to learn
the simple-to-complex wiring from their view-specific to
view-tolerant layers. They associate temporally adjacent frames from
the video of visual experience as in, e.g., \citep{Isik2012}.

The computational insight enabling depth-rotation tolerant
representations to be learned from experience is that, due to
properties of how objects move in the world, temporally adjacent
frames (the $g_i w^k$) almost always depict the same object
\citep{hinton1990unsupervised, stryker1991temporal, Foldiak1991,
  wiskott2002slow, berkes2009structured, Isik2012}. Short videos
containing a face almost always contain multiple views of the same
face. There is considerable evidence from physiology and psychophysics
that the brain employs a temporal-association strategy of this
sort \citep{miyashita1988neuronal, Wallis2001, Cox2005, Li2008,
  wallis2009learning, Li2010}. Thus, our assumption here is that in
order to get invariance to non-affine transformations (like rotation
in depth), it is necessary to have a learning rule that takes
  advantage of the temporal coherence of object identity.

More formally, this procedure achieves depth-rotation tolerance because the set of rotations in depth approximates the group structure of affine transformations in the plane (see Appendix section 1). For the latter case, there are theorems guaranteeing invariance without loss of selectivity by operations resembling the convolution in space performed by simple cells and the pooling done by complex cells \citep{anselmi2015invariance}.

Furthermore, \citep{leibo2015invariance} showed that Eq. \ref{signature} is approximately invariant to rotations in depth for $x$ a face, provided the templates $w^{k}$ also correspond to images of faces. For each template $w^k$, the rotated views $\{g_i w^{k}, ~~ i=1,\cdots, |G|\}$ must have been observed and stored. The $\eta (\scal{x}{g_i w^k})$ can be interpreted as the output of ``simple'' cells each with tuning ${g_i w^k}$ when stimulated with image $x$. In a similar way $\mu^k(x)$ can be interpreted as the activity of the ``complex'' cell indexed by $k$.

\subsection*{Biologically plausible learning}

The simple-complex algorithm described above can provide an invariant
representation but relies on a biologically implausible learning step:
storing a set of discrete views observed during
development. Instead we propose a more biologically plausible
mechanism: Hebb-like learning \citep{Hebb1949} at the level of simple
cells (see Equation \eqref{eq:oja_rule}). Instead of storing separate
frames, cortical neurons exposed to the rotation in depth of a face
update their synaptic weights according to a Hebb-like rule,
effectively becoming each tuned to one of a set of basis functions
corresponding to different combinations of the set of views. Different
Hebb-like rules lead to different sets of basis functions such as
Independent Components (IC) or Principal Components (PC). Since each of the neurons become tuned to one of these basis
functions instead of one of the views, a set of basis functions
replaces the ${g_i w^k}$ (for a given $k$) in the pooling Equation \eqref{signature}. The question is whether
invariance is still present under this new tuning.

The surprising answer is that most unsupervised learning rules will learn approximate invariance to viewpoint when provided with the appropriate
training set (see Appendix section 2 for a proof). In fact, unsupervised Hebb-like plasticity rules such as Oja's, Foldiak's trace rule, and ICA provide a basis that when used in the pooling equation provide invariance. Supervised learning rules such as backpropagation also satisfy the requirement as long as the training set is appropriate.

In the following we consider as an example a simple Hebbian learning scheme called Oja's rule \citep{oja1982simplified, oja1992principal}. At this point we are concerned only with establishing the model and why it computes a view-tolerant face representation. For this purpose we could use any of the other learning rules---like Foldiak's trace rule or ICA---but we focus on the Oja rule because it will turn out to be of singular relevance for mirror symmetry.  

The Oja rule can be derived as the first order expansion of a normalized Hebb rule. The assumption of this normalization is plausible, because normalization
mechanisms are widespread in cortex \citep{Turrigiano2004}.

For learning rate $\alpha$, Oja's rule is
\begin{equation}\label{eq:oja_rule}
	\Delta w = \alpha (xy - y^2w) = \alpha(xx^\intercal w - \left(w^\intercal x x^\intercal w) w\right).
\end{equation}

The original paper of Oja showed that the weights of a neuron updated
according to this rule will converge to the top principal component
(PC) of the neuron's past inputs, that is to an eigenvector of the
input's covariance $C$. Thus the synaptic weights correspond to the
solution of the eigenvector-eigenvalue equation $Cw = \lambda
w$. Plausible modifications of the rule---involving added noise or
inhibitory connections with similar neurons---yield additional
eigenvectors \citep{sanger1989optimal, oja1992principal}.  This
generalized Oja rule can be regarded as an online algorithm to compute
the principal components of incoming stream of vectors, in our case, images.

What is learned and how it is stored depends on the choice of a timescale over which learning takes place since learning is dictated by the underlying covariance $C$ of the inputs (see Appendix, section 3). In order for familiar faces to be stored so that the neural response modeled by Eq. \eqref{signature} tolerates rotations in depth of novel faces, we propose that Oja-type plasticity leads to representations for which the $w_i^k$ are given by principal components (PCs) of an image sequence depicting depth-rotation of face $k$. Consider an immature functional unit exposed, while in a plastic state, to all depth-rotations of a face. Oja's rule will converge to the eigenvectors corresponding to the top $r$ eigenvalues and thus to the subspace spanned by them.  The Appendix, section 2 shows that for each template face $k$ the signature $\mu^k(x) = \sum_{i = 1}^{r}\eta(\scal{x}{w_i^k})$ obtained by pooling over all PCs represented by different $w_i^k$ is an invariant. This is analogous to Eq. \eqref{signature} with $g_i w^k$ replaced by the $i$-th PC. The appendix also shows that other learning rules for which the solutions are not PCs but a different set of basis functions, generate invariance as well---for instance, independent components (see Appendix section 2).

\subsection*{Empirical evaluation of view-invariant face recognition performance}

\begin{figure}
\begin{center}
\centerline{\includegraphics[width=0.7\textwidth]{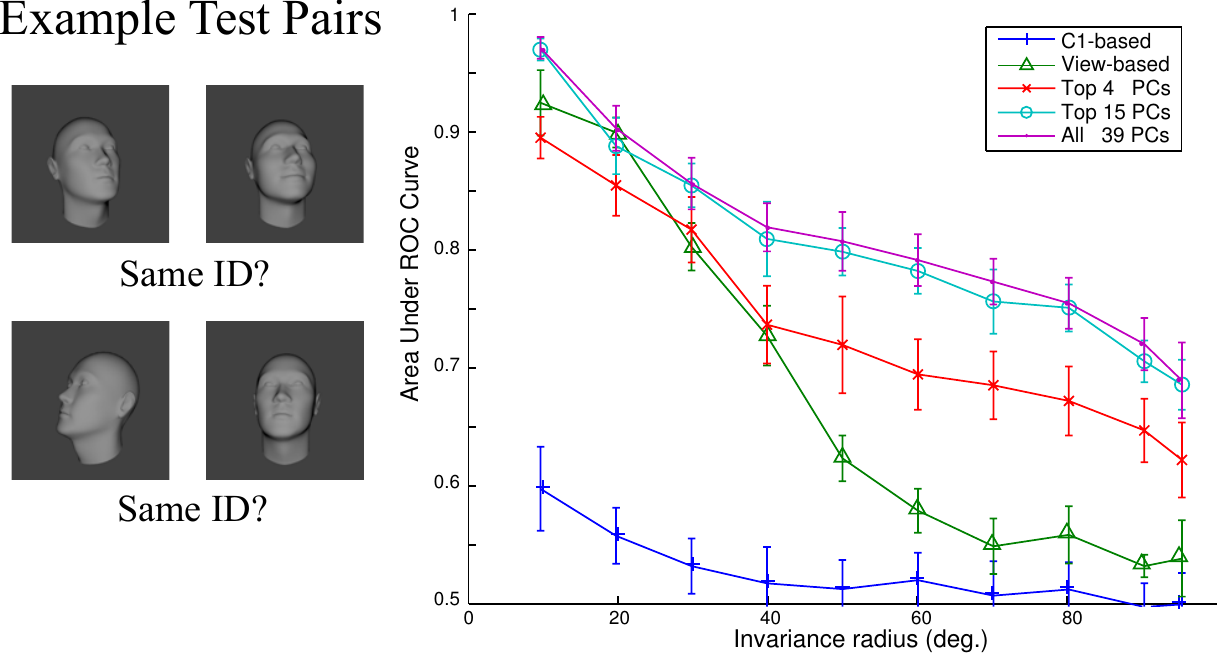}}   
\caption{\label{fig:performance} Model performance at the task of
  same-different pair matching as a function of the extent of depth
  rotations appearing in the test set (the invariance range of the
  task). All models were based on HMAX C1 features \citep{Serre2007a}.}
\end{center}
\end{figure}

View-invariance of the two models was assessed by simulating a
sequence of same-different pair-matching tasks, each demanding more
invariance than the last. In each test, 600 pairs of face images were
sampled from the set of faces with orientations in the current testing
interval. 300 pairs depicted the same individual and 300 pairs
depicted different individuals. Testing intervals were ordered by
inclusion and were always symmetric about $0^\circ$, the set of
frontal faces; i.e., they were $[-r, r]$ for $r = 5^\circ,\dots,
95^\circ$. The radius of the testing interval $r$, dubbed the
invariance range, is the abscissa in Fig. \ref{fig:performance}.

To classify an image pair $(a,b)$ as depicting the same or a different individual, the cosine similarity $(a\cdot b) / (\|a\| \|b\|)$ of the two representations was  compared to a threshold. The threshold was varied systematically in order to compute the area under the ROC curve (AUC), reported on the ordinate of Fig. \ref{fig:performance}. AUC declines as the range of testing orientations is widened. As long as enough PCs are used, the proposed model performs on par with the view-based model. It even exceeds its performance if the complete set of PCs is used. Both models outperform the baseline HMAX C1 representation (Fig. \ref{fig:performance}).

\subsection*{Mirror symmetry}

\begin{figure}
\begin{center}
\centerline{\includegraphics[width=0.65\textwidth]{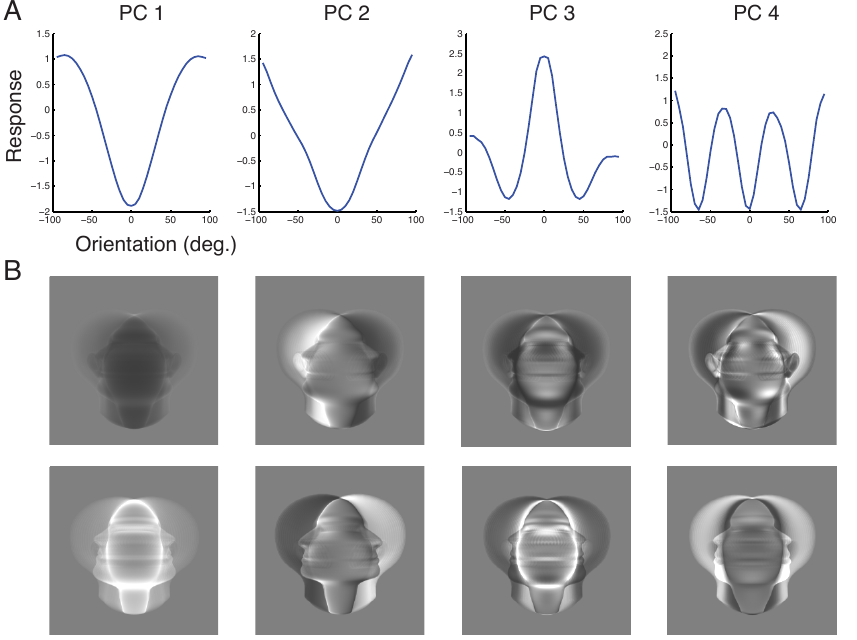}}    
\caption{Mirror symmetric orientation tuning of the raw pixels-based
  model (A) $(w_i \cdot x_\theta)^2$ as a function of the orientation
  of $x_\theta$. Here each curve represents a different PC. (B)
  Solutions to the Oja equation ($w_i$) visualized as images. They are
  either symmetric or antisymmetric about the vertical
  midline. \label{fig:orientation_tuning}}
\end{center}
\end{figure}

Consider the the case where, for each of the templates $w^k$, the developing organism has been exposed to a sequence of images showing a single face rotating from a left profile to a right profile. Faces are approximately bilaterally symmetric. Thus, for each face view $g_i w^k$, its reflection over the vertical midline $g_{-i} w^k$ will also be in the training set. It turns out that this property---along with the assumption of Oja plasticity, but not other kinds of plasticity---is sufficient to explain mirror symmetric tuning curves. The argument is as follows.

\noindent Consider a face, $x$ and its orbit in $3D$ w.r.t. the rotation group:
\begin{equation*}
O_x=(r_{0}x,\cdots,r_{N}x).
\end{equation*}
where $r$ is a rotation matrix in 3D, w.r.t., e.g., the $z$ axis.\\

\noindent Projecting onto $2D$ we have
\begin{equation*}
P(O_x)=(P(r_{0}x),\cdots,P(r_{N}x)).
\end{equation*}

\noindent Note now that, due to the bilateral symmetry, the above set can be written as:
\begin{equation*}
P(O_x)=(x_0,\cdots,x_{\frac{N}{2}},Rx_1,\cdots,Rx_{\frac{N}{2}}).
\end{equation*}
where $x_{n}=Pr_{n}x$, $n=1,\cdots,N/2$ and $R$ is the reflection
operator. Thus the set consists of a collection of orbits w.r.t. the
group $G=\{e,R\}$ of the templates $\{x_{1},\cdots,x_{N/2}\}$.  

This property of the training set is used in the appendix in two ways. First, it is needed in order to show that the signature $\mu(x)$ computed by pooling over the solutions to any equivariant learning rule, e.g., Hebb, Oja, Foldiak, ICA, or supervised backpropagation learning, is approximately invariant to depth-rotation (sections 1 -- 2).  

Second, in the specific case of the Oja learning rule, it is this same property of the training set that is used to prove that the solutions for the weights (i.e., the PCs) are either even or odd (section 3). This in turn implies that the penultimate stage of the signature computation: the stage where $\eta(\scal{w,x})$ is computed, will have orientation tuning curves that are either even or odd functions of the view angle. 

Finally, to get mirror symmetric tuning curves like those in AL, we need one final assumption: the nonlinearity before pooling at the level of the ``simple''  cells in AL must be an even nonlinearity such as $\eta(z) = z^2$. This is the same assumption as in the "energy model'' of \citep{adelson1985spatiotemporal}. This assumption is needed in order to predict mirror symmetric tuning curves for the neurons corresponding to odd solutions to the Oja equation. The neurons corresponding to even solutions have mirror symmetric tuning curves regardless of whether $\eta$ is even or odd.

\begin{figure}
\begin{center}
\centerline{\includegraphics[width=0.65\textwidth]{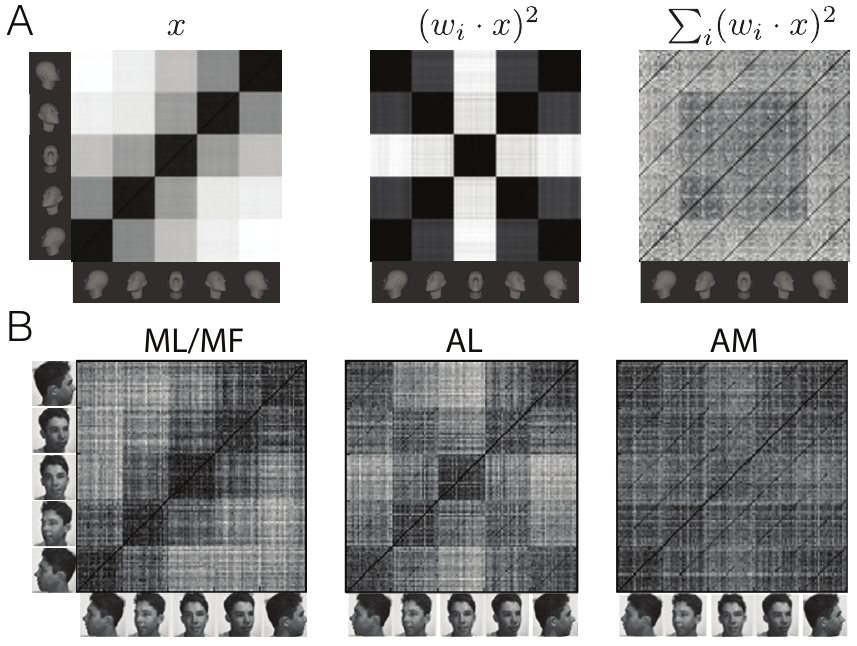}} 
\caption{Population representations of face view and identity (A) Model population similarity matrices (B) Neural population similarity matrices from \citep{freiwald2010functional}.\label{fig:correlation_matrices}}
\end{center}
\end{figure}

An orientation tuning curve is obtained by varying the orientation of the test image $\theta$. Fig. \ref{fig:orientation_tuning}-A shows example orientation tuning curves for the model based on a raw pixel representation. It plots $(\scal{x_\theta}{w_i})^2$ as a function of the test face's orientation for five example units tuned to features with different corresponding eigenvalues. All of these tuning curves are symmetric about $0^\circ$---i.e., the frontal face orientation. Fig. \ref{fig:correlation_matrices}-A shows how the three populations in the C1-based model represent face view and identity and Fig. \ref{fig:correlation_matrices}-B shows the same for populations of neurons recorded in ML/MF, AL, and AM. The model is the same one as in Fig. \ref{fig:performance}.

In contrast to the Oja/PCA case, we show through a simulation analogous to Fig. \ref{fig:correlation_matrices} that ICA does not yield mirror symmetric tuning curves (appendix section 4). Though this is an empirical finding for a specific form of ICA, we do not expect, based on our proof technique for the Oja case, that a generic learning rule would predict mirror symmetric tuning curves. 

These results imply that if neurons in AL learn according to a broad class of Hebb-like rules, then there will be invariance to viewpoint. Different AM cells would come to represent components of a view-invariant signature---one per neuron. Each component can correspond to a single face or to a set of faces, different for each component of the signature. Additionally, if the learning rule is of the Oja-type and the output nonlinearity is, at least roughly, squaring, then the model predicts that on the way to view invariance, mirror-symmetric tuning  emerges, as a necessary consequence of the intrinsic bilateral symmetry of faces.

\section*{Discussion}

The model discussed here provides a computational account of how
experience and evolution may wire up the ventral stream circuitry to
achieve the computational goal of view-invariant face recognition. 
Neurons in top-level face patch AM maintain an explicit representation
selective for face identity and tolerant to position, scale, and
viewing angle \citep{freiwald2010functional} (along with other units tolerant
to identity but selective for other variables such as viewing angle). The approach in this
paper explains how this property may arise in a feed-forward
hierarchy. To the best of our knowledge, it is the first account that
provides a computational explanation of why cells in the face
network's penultimate processing stage, AL, are tuned symmetrically to
head orientation. 

Our assumptions about the architecture for invariance conform to
i-theory \citep{anselmi2015unsupervised, anselmi2015invariance} which
is a theory of invariant recognition that characterizes and generalizes the convolutional and
pooling layers in deep networks. i-theory has recently been shown to
predict domain-specific regions in cortex
\citep{leibo2015invariance} with the function of achieving invariance
to class-specific transformations (e.g. for faces) and the specific form of
eccentricity-dependent cortical magnification
\citep{poggio2014computational}. Our assumption of Hebbian-like
plasticity for learning template views is, however, outside the
mathematics of i-theory: it links it to biological properties of
cortical synapses.

This argument of this paper has been made, as nearly as possible, from first principles. It begins with a claim about the computational problem faced by a part of the brain: the need to compute view-tolerant representations for faces. Yet it seeks to explain properties of single neurons in a specific brain region, AL, far from the sensory periphery.  The argument proceeds by considering which of the various biologically-plausible learning rules satisfy requirements coming from the theory while also yielding non-trivial  predictions for AL neurons in qualitative accord with the available data. It seems significant then that the argument only works in the case of Oja-like plasticity; it may suggest the hypothesis that such plasticity may indeed be driving learning in AL. 

The class of learning rules yielding invariance includes those that emerge from principles such as sparsity and the efficient coding hypothesis  \citep{attneave1954some, barlow1961possible, Olshausen1996}. However, explaining the mirror symmetric tuning of AL neurons apparently requires the Oja rule. An interesting direction for future work in this area could be to investigate the role of sparsity in the face processing system. Perhaps a learning algorithm derived from the efficient coding perspective that also explains AL's mirror symmetry could be found.

Our model is designed to account only for the feed-forward processing
in the ventral stream. Back-projections between visual areas---and of
course within each area---are well known to exist in the ventral
stream and probably also exist in the face patch network. They are
likely to play a major role in visual recognition after
$\sim80$ ms from image onset. Representations computed in the first
feeforward sweep are likely used to provide information about a
few basic questions such as the identity or pose of a face.
Additional processing is likely to require iterations and even
top-down computations involving shifts of fixation and generative
models. An example for face recognition is recent work
\citep{Yildirim2015} which combines a feedforward network like
ours---also showing mirror-symmetric tuning of cell populations---with
a probabilistic generative model. Thus our feedforward model, which
succeeds in explaining the main tuning and invariance properties of
the macaque face-processing system, may serve as a building block for
future object-recognition models addressing brain areas such as
prefrontal cortex, hippocampus and superior colliculus, integrating
feed-forward processing with subsequent computational steps that
involve eye-movements and their planning, together with task
dependency and interactions with memory.

\section*{Materials} 
\subsection*{Stimuli}
40 face models were rendered with perspective projection. Each face
was rendered (using Blender \citep{Stichting_Blender_Foundation}) at
each orientation in $5^\circ$ increments from $-95^\circ$ to
$95^\circ$. The untextured face models were generated using Facegen
\citep{Singular_Inversions}. All faces appeared on a uniform gray
background.

\subsection*{View-invariant Same-different Pair Matching Task}
For each of the 5 repetitions of the same-different pair matching
task, 20 template and 20 test individuals were randomly selected from
the full set of 40 individuals. The template and test sets were chosen
independently and were always disjoint. 50\% of the 600 test pairs
sampled from each testing interval depicted the same two
individuals. Each testing interval was symmetric about $0^\circ$
(frontal) and testing intervals were ordered by inclusion. The
smallest was $[-10^\circ, 10^\circ]$ and the largest was $[-95^\circ,
95^\circ]$ (left and right profile views). The classifier compared the
Cosine similarity of the two zero-mean, and unit-standard deviation
representations to a threshold. The threshold was integrated over to
compute the area under the ROC curve (AUC). The abscissa of
Fig. \ref{fig:performance} is the radius of the testing interval from
which test pairs were sampled. The ordinate of
Fig. \ref{fig:performance} is the mean AUC $\pm$ the standard
deviation computed over the 5 repetitions of the experiment.

A similarity matrix in Figure \ref{fig:correlation_matrices} was obtained by computing Pearson's linear
correlation coefficient between each test sample pair. The same matrix
was computed 10 times with different training/test splits and the
average was reported. Same procedures were repeated for features from
area MLMF, AL and AM to get corresponding matrices.

\section*{Acknowledgments}
This material is based upon work supported by the Center for Brains,
Minds, and Machines (CBMM), funded by NSF STC award CCF-1231216.  This
research was also sponsored by grants from the National Science
Foundation (NSF-0640097, NSF-0827427), and AFOSR-THRL
(FA8650-05-C-7262). Additional support was provided by the Eugene
McDermott Foundation.
 
\bibliographystyle{plainnat}
\bibliography{biblio}

\clearpage     
  
\appendix

\section*{Appendix}

The key results in this appendix can be informally stated as follows:

\begin{itemize}
\item We prove than a number of learning rules, supervised and
  unsupervised, are equivariant with respect to the symmetries of the
  training data. We use this result in the case of training data consisting
  of images of faces for all view angles obtaining equivariance of the
  solutions of the learning rules with respect to the reflection group
  and the group of rotations. The implications that we use in
  the paper are

\begin{itemize}

\item  the solutions of all learning rules can be used as templates in
  the computation of an invariant signature. The algorithm consists of performing dot products
  of the input image with each template, transforming nonlinearly (for
  instance using a rectifier nonlinearity or a square) the result and then
  pooling over {\it all} templates, i.e., the solutions of the
  learning rule. The result is approximately invariant to rotation in
  depth.

\item in the case of the Oja rule we prove that the solutions are even or odd
  functions of the view angle; a square nonlinearity provides even
  functions, which are mirror-symmetric. We were not able to prove
  such a property for any of the other learning rules.
\end{itemize}

\item in the case of the ICA rule we show empirical evidence that the
  solutions are neither odd nor even. This suggests that most learning
  rules do not lead to even or odd solutions.
\end{itemize}

The appendix is divided into four sections:

\begin{enumerate}
\item In section \ref{ApproximateGroup} we show how recent theorems on
  invariance under group transformations could be extended to
  nongroups and under which conditions.  We show
  how an {\it approximately invariant} signature can be computed in
  this setting. In particular we analyze the case of rotation in depth
  and mirror symmetry transformations of bilateral symmetric objects
  such as faces.
\item In section \ref{Equivariance} we describe how the group symmetry properties of
  the set of images to which neurons are exposed (the ``unsupervised''
  training set) determine the symmetries of the learned weights. In
  particular we show how the weight symmetries gives a simple way of
  computing an invariant signature.

\item In section \ref{Oja} we prove that the solutions of the Oja equation, given that the input
  vectors that are reflections of each other (like a face's view at
  $\theta$ degrees and its view at $-\theta$ degrees), must be odd or even.

\item In section \ref{ICA} we provide empirical evidence that there are
  solutions of ICA algorithms---on the same data as above---that do
  not show any symmetry.

\end{enumerate}
\noindent
In the following we indicate with $x\in R^d$ an image, with $w\in R^d$ a filter or neural weight and with $G$ a locally compact group.

\section{Approximate Invariance for non-group transformations}
\label{ApproximateGroup}
In this section we analyze the problem of getting an approximately invariant signature for image transformations that do not have a group structure.  In fact, clearly, not all image transformations have a group structure. However
assuming that the object transformation defines a smooth manifold
we have (by the theory of Lie manifolds) that locally a Lie group is
defined by the generators on the tangent space.
\noindent
We illustrate this in a simple example.
Let $x\in \R^d$. Let $s:\R^d \times \R^{Q} \rightarrow \R^d$ a $C^{\infty}$ transformation depending on $\Theta=(\theta_{1},\cdots,\theta_{Q})$ parameters.
For any fixed $x\in \R^d$ the set $M =(s(x,\Theta),\;\Theta\in \R^{Q})$ describe a differentiable manifold.
If we expand the transformation around e.g. $\vec{0}$ we have:
\begin{equation}\label{approx}
s(x,\Theta) = s(x,\vec{0}) + \sum_{i=1}^{Q}\frac{\partial s(x,\Theta)}{\partial \theta_{i}}\theta_{i} + o(\nor{\Theta}^{2}) = x + \sum_{i=1}^{Q}\theta_{i} L_{\theta_{i}}(x) + o(\nor{\Theta}^{2}) \end{equation}
where $L_{\theta_{i}}$ are the infinitesimal generators of the transformation in the $i^{th}$ direction.\\
Therefore locally (when the term $o(\nor{\Theta}^{2})$ can be neglected) the associated group transformation can be expressed by exponentiation as:
$$
g(\Theta) = \exp
(\theta_{1}L_{\theta_{1}}+\theta_{2}L_{\theta_{2}}+\cdots+\theta_{Q}L_{\theta_{Q}}).
$$
Note that the above expansion is valid only locally.  In other
words instead of a global group structure of the transformation we
will have a collection of local transformations that obey a group
structure. The results derived
in section \ref{Equivariance} will then say that the local learned weights will be
orbits w.r.t. the local group approximating the non-group global
transformation.

\subsection{Invariance under rotations in depth}
\label{Invariancerotindepth}
The 3D ``views'' of an object undergoing a 3D
rotation are group transformations but the 2D projections of an object undergoing a 3D
rotation are not group transformations. However for any fixed angle
$\theta_{0}$ and for small rotations the projected images
approximately follow a group structure. This can be easily seen making
the substitution in eq. \eqref{approx} $s(x,\Theta)=P(r_{\theta}x)$
where $P$ is the 2D projection. Let $\eta: \R \rightarrow \R$ be a nonlinear function, e.g., squaring or rectification. For small values of $\theta$ we have
therefore that the signature:
$$
\mu_{w}(x)=\int^{\theta_{0}}_{-\theta_{0}}\;d\theta\;\eta(\scal{Px}{Pr_{\theta}w})
$$
or its discrete version
$$
\mu_{w}(x)=\sum_{i}\;\eta(\scal{Px}{Pr_{\theta_{i}}w})=\sum_{i}\;\eta(\scal{Px}{g(\theta_{i})Pw})
$$
is invariant under 3D rotation of $x$ of an angle $\bar{\theta}$ up to
a factor proportional to $O(\nor{\bar{\theta}})$.  Alternatively if
the following property holds:
\begin{equation}\label{conc}
\scal{Px}{Pr_{\theta}w}=0\;\;\theta>\bar{\theta}
\end{equation}
the invariance will be exact (see \cite{Anselmi2013, leibo2015invariance}); this is the case e.g. when both $w$ and $x$ are faces.

The locality of the group structure (eq. \eqref{conc}) means that we have invariance of
the signature only within each local neighborhood but not over all
viewpoints. A reasonable scenario could be that each local neighborhood may consist of, say, $\pm 30$
degrees (depending on the universe of distractors). Almost complete
view invariance can be obtained from a single view at $+30$
degrees. In fact the view, together with the associated virtual view
at $-30$ degrees because of mirror symmetry,
provides invariance over $-60,+60$ degrees \cite{poggio19923d}.

\subsection{Rotation in depth and mirror symmetry.}\label{Mirrorsymmetryorbits}
As explained on the previous paragraph, projected rotations in depth are not group transformations.
However in the case of a bilateral symmetric objects, as we will see below, projected rotations in depth are a collection of orbits of the mirror symmetry group. Section \ref{Equivariance} will clarify why this property is important proving that it forces the set of solutions of a variety of learning rules to be a collection of orbits w.r.t. the mirror symmetry group.\\
\noindent
Consider e.g. a face, $x$, which is a bilateral symmetric object and its orbit in $3D$ w.r.t. the rotation group:
$$
O_x=(r_{0}x,\cdots,r_{2\pi}x).
$$
where $r$ is a rotation matrix in 3D, e.g. w.r.t. the $z$ axis.\\
\noindent
Projecting onto $2D$ we have
$$
P(O_x)=(P(r_{0}x),\cdots,P(r_{2\pi}x)).
$$
Note now that, due to the bilateral symmetry, the above set can be written as:
$$
P(O_x)=(x_0,\cdots,x_{\frac{N}{2}},Rx_1,\cdots,Rx_{\frac{N}{2}}).
$$
where $x_{n}=Pr_{\theta_{n}}x$, $n=1,\cdots,N/2$ and $R$ is the reflection operator. The set consists of a collection of orbits w.r.t. the group $G=\{e,R\}$.   This is due to the relation
$$
x_n=P(r_{\theta_{n}}x)=Rx_{\frac{N}{2}+n}=R(Pr_{-\theta}x).
$$
i.e. a face rotated by an angle $\theta$ and then projected is equal to the reflection of the same face rotated by an angle $-\theta$ and projected.\\
\noindent
The reasoning generalizes to multiple faces. In summary  in the specific case of bilateral symmetric objects rotating in depth,  a projection onto a plane  parallel to the rotation axis creates images which are transformations w.r.t. the group of reflection, thus falling in the group case described in the above paragraphs.

\section{Unsupervised and supervised learning and data symmetries}
\label{Equivariance}
In the following we show how symmetry properties on the neuronal inputs affect the learned weights.
We model different unsupervised (Hebbian, Oja, Foldiak, ICA) or
supervised learning (SGD) rules as dynamical systems coming from the requirement of minimization of some target function. We see how these dynamical systems are equivariant (in the sense specified below)
and how equivariance determines the symmetry properties of their solutions.

This gives a simple way to generate an invariant signature by averaging over all solutions.

\subsection{Equivariant dynamical systems and their solutions.}\label{dynamicalsystems}
We make the general assumption that the dynamical system can be
described in terms of trying to minimize a non-linear functional of the form:
\begin{equation}\label{general}
\argmin_{w\in X}\;\;\mathcal{L}(w,x),\;\;\;\;\mathcal{L}(w,x) = h(w,x),\;\;\;\;x,w\in\R^d
\end{equation}
The associated dynamical system reads as:
\begin{equation}\label{GD}
\dot{w}=f(w)=\dot{h}(w,x).
\end{equation}
A general result holds for equivariant dynamical systems. A dynamical system is called \textit{equivariant} w.r.t. a group $G$ if $f$ in eq. \eqref{GD} commutes with any transformation $g\in G$ i.e.
\begin{equation}\label{dsystem}
 f(gw)=gf(w),\;\;\;\forall g\in G.
\end{equation}
In this case we have:\\
\begin{theorem}
If an equivariant dynamical system has a solution $w$, then the whole group orbit of $w$ will also be a set of solutions (see \cite{golu}).
\end{theorem}
In the following we are going to analyze different cases of updating rules for neuronal weights showing, under the hypothesis that the training set is a (scrambled) collection of the orbits  i.e. we specialize the set $X$ to be of the form:
\begin{equation}\label{hp}
X=G\mathcal{T},\;\;\T\in\R^{d\times N},\;\;X=\{x_{1},\cdots,x_{N}\},
\end{equation}
that the dynamical system is equivariant.\\
\noindent
We will see that the following variant of the equivariance holds for many dynamical systems:
\begin{equation}\label{dsystemp}
 f(gw,x)=gf(w,\pi_{g}(x)),\;\;\;\forall g\in G,\; x\in X.
\end{equation}
where $\pi_{g}(x)$ is permutation of the set $X$ that depends on $g$.
The derivation stands on the simple observation:
$$
\scal{x}{gw}= \scal{g^{-1}x}{w}
$$
and the hypothesis that the training set is a collection of orbits. In fact in this case
$$
gX=\pi_{g}(X).
$$
In general if the training set $X$ is large enough the dynamical
system will be equivalent to the unpermuted one due to the stability
of the stochastic gradient descent method \cite{hardt}.
Since the dynamical systems associated with the Oja and the  ICA rules minimize statistical moments they are clearly independent of training data permutations. The fact that the set of solutions is a collections of orbits,   $S=\bigcup_{i}O_{i}$ implies that any average operator over them is invariant. In our case the operator is the signature:
$$
\mu(x)= \sum_{ij}\eta(\scal{x}{O_{ij}})
$$
where $O_{ij}$ is the element $j$ of the orbit $i$ and $\eta:\R\to\R$ is a non-linear function.\\\\
\noindent
In the following we prove equivariance of a few learning rules.

\begin{enumerate}

\item{\textbf{Unsupervised learning} rules\cite{hassoun}}:

In the following $x\in X$ and $\alpha>0$ and with the notation $\pi_g(x)$ we indicate the permutation of the element $x$ in the training set $X$ due to the transformation $g$.
\begin{itemize}

\item{\textbf{Hebbian learning}.}
Choosing
\begin{equation}
\mathcal{L}(w,x)=\frac{\alpha}{2}y^2
\end{equation}
where $y=\scal{x}{w}$ is the neuron's response, we have the associated dynamical system is:
\begin{equation}
\dot{w}=f(x,w)=\alpha\scal{x}{w}x.
\end{equation}
The system is equivariant. In fact:
\begin{equation*}
f(x,gw)= \alpha\scal{x}{gw}x=g\alpha\scal{g^{-1}x}{w}g^{-1}x=g\alpha\scal{\pi_{g}(x)}{w}\pi_{g}(x)=gf(\pi_{g}(x),w).
\end{equation*}

\item{\textbf{Oja learning}.}
Choosing
\begin{equation}
\mathcal{L}(w,x)=\frac{\alpha}{2\nor{w}_{2}}\scal{x}{w}^2
\end{equation}
we obtain by differentiation:
\begin{equation}
\dot{w}= f(w,x)=\alpha \frac{y}{\nor{w}_{2}}(x-y\frac{w}{\nor{w}_{2}}).
\end{equation}
The obtained dynamical system is that of Oja's for the choice $\nor{w}_{2}=1$.
The system is equivariant (note that $\nor{gw}_{2}=\nor{w}_{2}$). In fact:
\begin{eqnarray*}
f(gw,x)&=& \alpha\scal{x}{gw}(x-\scal{x}{gw}gw)=\alpha\scal{g^{-1}x}{w}g(g^{-1}x-\scal{g^{-1}x}{w}w)\\
&=&\alpha\scal{\pi_{g}(x)}{w}g(\pi_{g}(x)-\scal{\pi_{g}(x)}{w}w)=gf(w,\pi_{g}(x))
\end{eqnarray*}

\item {\textbf{ICA}}. Choosing
\begin{equation}
\mathcal{L}(w,x)=\alpha\frac{\scal{x}{w}^{4}}{4}+\frac{\nor{w}_{2}^{2}}{2}
\end{equation}
we obtain the dynamical system:
\begin{equation}
\dot{w}= \alpha(\scal{x}{w}^3x-w)
\end{equation}
which can be shown to extract one ICA component \cite{hyvarinen1998independent}. The system is equivariant. In fact:
\begin{equation*}
f(x,gw)=\alpha(\scal{x}{gw}^3x-gw) = g\alpha(\scal{g^{-1}x}{w}^3g^{-1}x -w) = gf(\pi_{g}(x),w).
\end{equation*}
\item {\textbf{Foldiak.}}
Choosing:
\begin{equation}
\mathcal{L}(x,w)=\frac{\alpha}{2} \bar{y}^2,\;\;\;\bar{y}=\int_{t_{0}}^{t}\;d\tau\;\scal{w}{x}(\tau)
\end{equation}
the associated dynamical system is:
\begin{equation}
\dot{w}(t) = \alpha\Big(\int_{t_{0}}^{t}d\tau\;\scal{w}{x}(\tau)\Big)x(t)=\alpha \bar{y}x(t)
\end{equation}
which is the so called Foldiak updating rule. The system is equivariant. In fact:
\begin{eqnarray*}
f(x,gw) &=& \alpha\Big(\int_{t_{0}}^{t}d\tau\;\scal{gw}{x}(\tau)\Big)x=g\alpha\Big(\int_{t_{0}}^{t}d\tau\;\scal{w}{g^{-1}x}(\tau)\Big)g^{-1}x\\
&=& \alpha g \bar{y}(w,\pi_{g}(x))\pi_{g}(x)=gf(w,\pi_{g}(x))
\end{eqnarray*}
\end{itemize}

\item{\textbf{Supervised learning in deep convolutional networks}.}
The reasoning above can be extended to supervised problems of the form:
\begin{equation}\label{general}
\argmin_{W}\;\; \mathcal{L}(X,\ell,W),\;\; X = (x_1, \ldots, x_N)
\end{equation}
where  $\mathcal{L}(X,\ell,W)=Loss(X,\ell,W)$.
The term $Loss(X,\ell,W)$ is a function defined using the loss of representing a set of observations $X$, their labels $\ell$, and a the set of the network weights $W$.
The updating rule for each weight $w_l$ is given by the backpropagation algorithm:
\begin{equation}
\dot{w}_{l}=\frac{\partial \mathcal{L}}{\partial w_{l}}.
\end{equation}
If the equation above is equivariant the same results of the previous section will hold, i.e., if there exists a solution the whole orbit will be a set of solutions.
In the following we analyze the case of deep networks showing that equivariance holds if the output at each layer $l$, $o_{l}$ is covariant w.r.t. the transformation, i.e.:
\begin{equation}\label{cov}
o_{l}(gx)=go_{l}(x),\;\;\forall\;g\in G
\end{equation}

We analyze the case of \textbf{deep convolutional networks} with pooling layers between each convolutional layer. In this case the response at each layer is covariant w.r.t. to the input transformation: the output at layer $l$ is  of the form:
\begin{equation}\label{layerl}
o_{l}(X,W_{l-1})(g)= \int_{gG_{l}}\;d\hat{g}\;\eta(\scal{o_{l-1}(X,W_{l-2})}{\hat{g}w_{l}}) = \int_{gG_{l}}\;d\hat{g}\;\eta(o_{l-1}(X,W_{l-2})*w_{l})(\hat{g})
\end{equation}
i.e. it is an average of a group convolution where $o_{l-1}$ is the output of layer $l-1$ and $W_{l-1}$ is the collection of weights up to layer $l-1$. Using the property that the group convolution commutes with group shift i.e.
$[(T_{\bar{g}}f)*h](g)=T_{\bar{g}}[f*h](g)$ we have:
\begin{eqnarray*}
o_{l}(\bar{g}X, W_{l-1})(g) &=& \int_{gG_{l}}\;d\hat{g}\;\eta(\bar{g}o_{l-1}(X,W_{l-2})*w_{l})(\hat{g})=\int_{gG_{l}}\;d\hat{g}\;\eta(o_{l-1}(X,W_{l-2})*w_{l})(\bar{g}\hat{g})\\
&=& \int_{\bar{g}gG_{l}}\;d\hat{g}\;\eta(o_{l-1}(X,W_{l-2})*w_{l})(\hat{g})= o_{l}(X,W_{l-1})(\bar{g}g)=\bar{g}o_{l}(X, W_{l-1})(g).
\end{eqnarray*}
where we used  the property $o_{l-1}(\bar{g}X,W)=\bar{g}o_{l-1}(X,W)$. This can be seen to hold using an inductive reasoning up to the first layer where:
$$
o_{2}(\bar{g}x, W_1)(g) = \int_{gG_{1}}\;d\hat{g}\;\eta((\bar{g}x)*w_{1})(\hat{g}) = \int_{\bar{g}gG_{1}}\;d\hat{g}\;\eta(x*w_{1})(\hat{g})=\bar{g}o_{1}(x,W_1)(g).
$$

In the following we prove that the dynamical systems (updating rules for the weights) associated to a deep convolutional network are equivariant. We consider e.g. the square loss function (the same reasoning can be extended to many commonly used loss functions):
$$
\mathcal{L}(\phi_L(X,W),\ell)=\sum_{\ell}(1-y_{\ell}\phi(X,W))^2.
$$
where 
$$
\phi_{L}(X,W)=\phi_{L}(\cdots,\phi_{3}(\phi_{2}(X,w_1),w_3),\cdots,w_{l}\cdots,w_L)
$$
being $L$ the layers number and $\ell$ is a set of labels. 
The associated dynamical system reads as:
$$
\frac{\partial \mathcal{L}(\phi_L(X,W),\ell)}{\partial w_{l}}= \dot{\mathcal{L}}(\phi_L(X,W),\ell)\frac{\partial \phi_{L}(X,W)}{\partial w_l}= 2\sum_{\ell}(1-y_{\ell}\phi_{L}(X,W))\frac{\partial \phi_{L}(X,W)}{\partial w_l}
$$
Substituting $w_l$ with $\bar{g}w_l$ we have, by the covariance property, that  the  first factor of the r.h.s. of the equation above becomes $\sum_{\ell}(1-y_{\ell}\phi_{L}(\pi_{\bar{g}}(X),W))$. We are then left to prove the equivariance of the second factor.\\
Using the chain rule, we have:
\begin{eqnarray*}
\dot{w}_{l}&=& \frac{\partial \phi_{L}(\cdots,\phi_{3}(\phi_{2}(x,w_1),w_2)\cdots,w_L) }{\partial w_l}\\
           &=& \dot{\phi}_{L}[o_{L}(W_{L-1},x)]\;\dot{\phi}_{L-1}[o_{L-1}(W_{L-2},x)]\;\cdots\;
           \dot{\phi}_{l}(o_{l-1}(x,W_{l-2}),w_l),\cdots,w_L)
\end{eqnarray*}
where $o_{j}(W_{l-1},x)=\phi_{j}(\cdots \phi_{l}(o_{l+1}(x,W_{l-1}),w_{l}),\cdots,w_L)$, $l<j<L$, being the output at layer $j$.\\
\noindent
Notice that, in the case of covariant layer outputs, we have:
\begin{eqnarray*}
\phi_{j}(\cdots \phi_{l+1}(o_{l}(X,W_{l-1}),\bar{g}w_l),\cdots,w_L)&=& \phi_{j}(\cdots \phi_{l+1}(\bar{g}^{-1}o_{l}(X,W_{l-1}),w_l),\cdots,w_L)\\
&=&\phi_{j}(\cdots \phi_{l+1}(o_{l}(\bar{g}^{-1}X,W_{l-1})),w_l),\cdots,w_L)\\
&=& \phi_{j}(\cdots \phi_{l+1}(o_{l}(\pi_{\bar{g}}(X),W_{l-1})),w_l),\cdots,w_L)
\end{eqnarray*}
where we used the covariance property in eq. \eqref{cov} and the fact that the training set is a collection of orbits w.r.t. the group $G$.\\
\noindent
Finally we have:
\begin{equation*}
\frac{\partial \mathcal{L}(X,\{w_1,\cdots,\bar{g}w_{l},\cdots,w_{L}\},\ell)}{\partial w_{l}}=\bar{g}\frac{\partial \mathcal{L}(\pi_g(X),\{w_1,\cdots,w_{l},\cdots,w_{L}\},\ell)}{\partial w_{l}}
\end{equation*}
where the $\bar{g}$ comes from the derivative of $\bar{g}w_l$ w.r.t. $w_l$.\\
Summarizing we have the following  result
\begin{theorem}
For $i = 1, \dots, L$, let $\phi_{i}:\R^{d_i}\to R^{d_{i+1}}$ depend on a set of weights $w_i$. Consider a deep convolutional network with output of the form
\begin{equation}\label{lf}
\phi_L(X,W)=\phi_{L}(\cdots,\phi_{2}(\phi_{1}(X,w_1),w_2),\cdots,w_{l})\cdots,w_L).
\end{equation}
and  a differentiable square loss $\mathcal{L}(\phi_{L}(X,W),\ell)$, being $\ell$ a set of labels.\\
If $X$ is a collection of orbits and and each $\phi_{i}$ is covariant, then the associated dynamical systems for each layer's weights' evolution in time
\begin{equation*}
\dot{w}_{l}= \frac{\partial \mathcal{L}(\phi_{L}(X,W),\ell)}{\partial w_{l}}
\end{equation*}
are equivariant w.r.t. the group $G$.
\end{theorem}
\end{enumerate}

\begin{figure}[t]
\begin{center}
\centerline{\includegraphics[width=17cm]{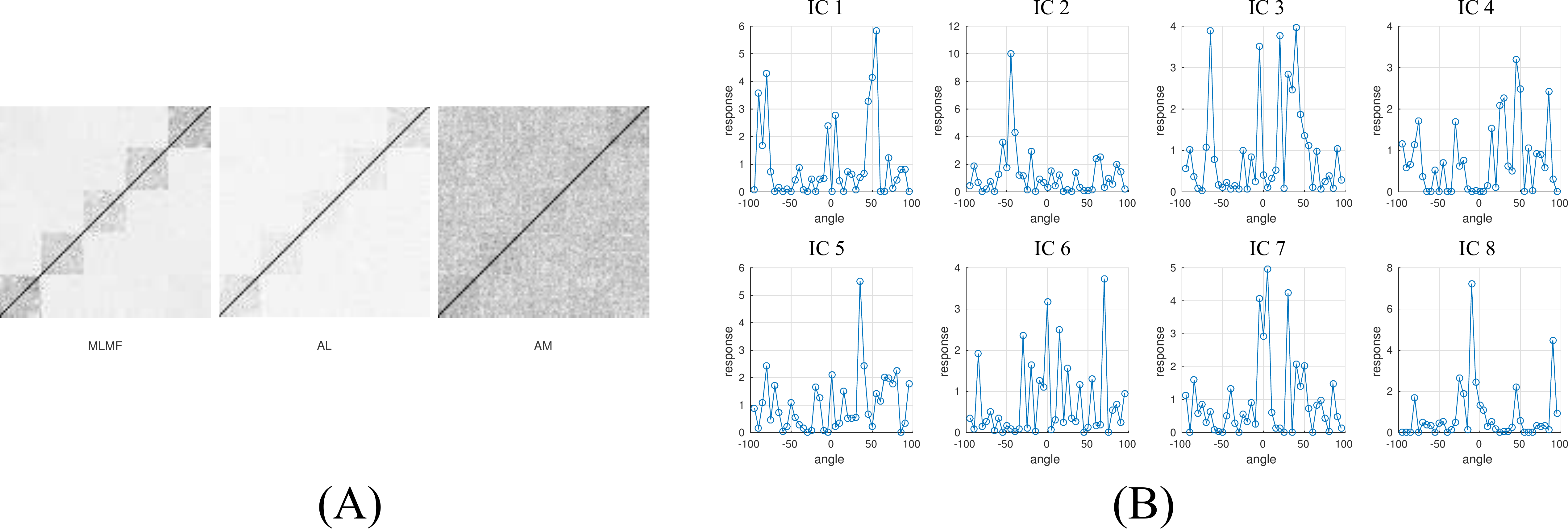}}
\caption{Experiments with ICA: we adopt the same pipeline as shown in
  the main text but replaced PCA with ICA \cite{hyvarinen1999fast}
  (which includes a ZCA-whitening preprocessing step). Similar to the
  original pipeline, for each training identity, ICA is performed to
  get 39 independent component directions. A testing image is
  projected to these directions. A square nonlinearity and a pooling
  are performed on the results. We show (A) the model population
  similarity matrices of different stages (similar to Fig 5A) and (B) some
  single cell responses in stage AL (similar to Fig 4A). Unlike PCA, the order of the independent components are arbitrary. \label{fig:ica}}
\end{center}
\end{figure}

\section{Proof that the Oja equation's solutions are odd or even.}\label{Oja}
So far we have shown how biologically plausible learning dynamics in conjunction with appropriate training sets lead to solutions capable of supporting the computation of a view-invariant face signature (Sections \ref{ApproximateGroup} -- \ref{Equivariance}). We showed that several different learning rules satisfied these requirements: Hebb, Oja, Foldiak, ICA, and supervised backpropagation (Section \ref{dynamicalsystems}). Now we use properties specific to the Oja rule to address the question of why mirror symmetric responses arise in an intermediate step along the brain's circuit for computing view-invariant face representations.

We now use the following well-known property of Oja's learning rule: that it implements an online algorithm for principal component extraction  \cite{oja1992principal}. More specifically, we use that the Oja dynamics converge to an eigenfunction of the training set's covariance $C(X)$.

Recall from section \ref{Mirrorsymmetryorbits} that in order to guarantee approximate view-invariance for bilaterally symmetric objects like faces, the training set $X$ must consist of a collection of orbits of faces w.r.t. the reflection group $G=(e,R)$. We now show that this implies the eigenfunctions of $C(X)$ (equivalently, the principal components (PCs) of $X$) must be odd or even.

Under this hypothesis the covariance matrix $C(X)$ can be written as
\begin{equation*}
C(X)=XX^\intercal= \T\T^\intercal + R\T\T^\intercal R^{\intercal}
\end{equation*}
where $\T$ is the set of the orbit representatives (untransformed vectors).\\
\noindent
It is immediate to see that the above implies $[C(X),R]=0$ (they commute). Thus $C(X)$ and $R$ must share the same eigenfunctions. Finally, since the eigenfunctions of the reflection operator $R$ are odd or even, this implies the eigenfunctions of $C(X)$ must also be odd or even.

Finally, we note that in the specific case of a frontal view, even basis functions (w.r.t. the zero view) are mirror symmetric.

\section{Empirical ICA solutions do not show any symmetry}\label{ICA}

Fig. \ref{fig:ica} shows results from the analogous experiment to main Fig. 4. but with ICA instead of PCA. Note that the ICA result is not mirror symmetric.

\end{document}

%% file: macros.tex
\usepackage{amsmath, amsthm}
\usepackage{amsfonts}
\usepackage{amssymb}

\newcommand{\R}{\mathbb{R}}

\DeclareMathOperator*{\argmin}{arg\,min}
